\documentclass[sigconf]{acmart}

\AtBeginDocument{%
  \providecommand\BibTeX{{%
    \normalfont B\kern-0.5em{\scshape i\kern-0.25em b}\kern-0.8em\TeX}}}

\copyrightyear{2024}
\acmYear{2024}
\setcopyright{rightsretained}
\acmConference[WWW'24 Companion]{Companion Proceedings of the ACM Web Conference 2024}{May 13--17, 2024}{Singapore, Singapore}
\acmBooktitle{Companion Proceedings of the ACM Web Conference 2024 (WWW '24 Companion), May 13--17, 2024, Singapore, Singapore}\acmDOI{10.1145/3589335.3651897}
\acmISBN{979-8-4007-0172-6/24/05}

\DeclareMathOperator{\Var}{Var}
\usepackage{amsmath}
\usepackage{multirow}

\usepackage{color,soul}

\begin{document}

\title{Towards Invariant Time Series Forecasting in Smart Cities}

\author{Ziyi Zhang}
\email{zyzhang@tamu.edu}
\affiliation{%
  \institution{Texas A\&M University}
  \city{College Station}
  \state{Texas}
  \country{USA}
}

\author{Shaogang Ren}
\email{shaogang@tamu.edu}
\affiliation{%
  \institution{Texas A\&M University}
  \city{College Station}
  \state{Texas}
  \country{USA}
}

\author{Xiaoning Qian}
\email{xqian@tamu.edu}
\affiliation{%
  \institution{Texas A\&M University}
  \city{College Station}
  \state{Texas}
  \country{USA}
}

\author{Nick Duffield}
\email{duffieldng@tamu.edu}
\affiliation{%
  \institution{Texas A\&M University}
  \city{College Station}
  \state{Texas}
  \country{USA}
}

\renewcommand{\shortauthors}{Ziyi Zhang, Shaogang Ren, Xiaoning Qian, \& Nick Duffield}

\begin{abstract} 
In the transformative landscape of smart cities, the integration of the cutting-edge web technologies into time series forecasting presents a pivotal opportunity to enhance urban planning, sustainability, and economic growth. The advancement of deep neural networks  has significantly improved forecasting performance. However, a notable challenge lies in the ability of these models to generalize well to out-of-distribution (OOD) time series data. The inherent spatial heterogeneity and domain shifts across urban environments create hurdles that prevent models from adapting and performing effectively in new urban environments. To tackle this problem, we propose a solution to derive invariant representations for more robust predictions under different urban environments instead of relying on spurious correlation across urban environments for better generalizability. Through extensive experiments on both synthetic and real-world data, we demonstrate that our proposed method outperforms traditional time series forecasting models when tackling domain shifts in changing urban environments. The effectiveness and robustness of our method can be extended to diverse fields including climate modeling, urban planning, and smart city resource management.
\end{abstract}


\maketitle

\section{Introduction}
Time series data plays a pivotal role in analyzing, monitoring, and simulating the development and design of smart cities. Extensive research across various domains has leveraged this data for applications including weather forecasting \cite{Weather}, temperature monitoring \cite{GTS}, and enhancing information systems \citep{TGNs}. Despite these advancements, analyzing time series data in urban environments introduces significant challenges due to geographic domain shifts. Such shifts represent a critical barrier in forecasting efforts, as models must not only capture temporal dependencies but also discern and adapt to invariant relationships within diverse and changing urban environments. This research seeks to address these challenges by developing a robust model capable of navigating the complexities introduced by urban variability, thereby contributing to the foundational technologies necessary for the smart cities of the future \citep{Reinforce}.

\begin{figure}[ht]
  \centering
  \includegraphics[width=\linewidth]{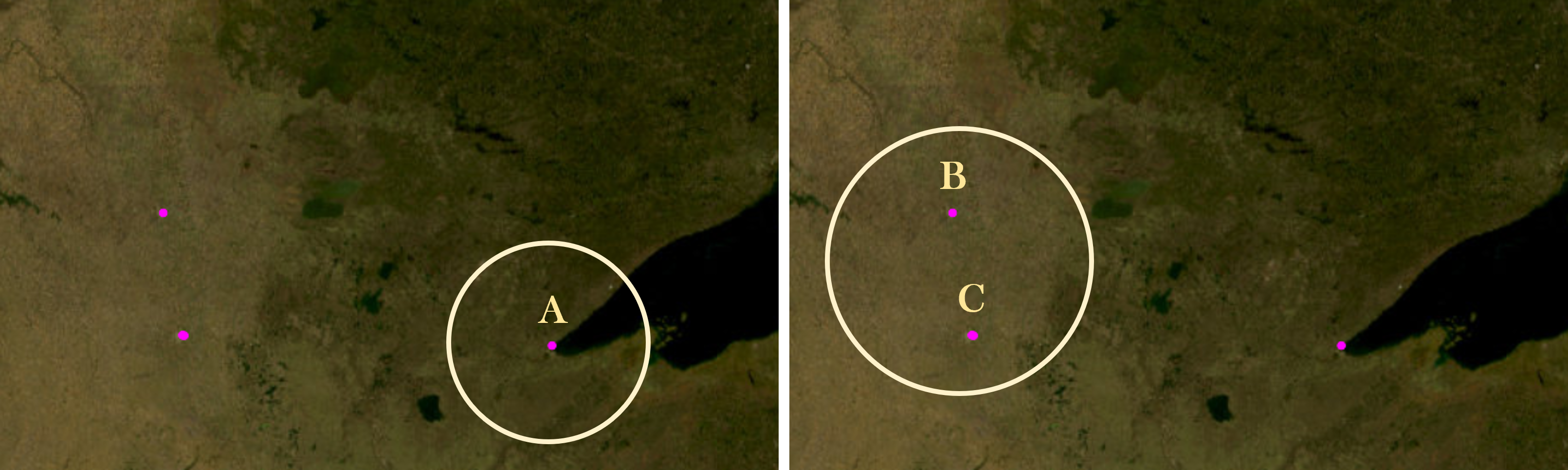}
  \captionsetup{justification=centering}
  \caption{Train a time series forecasting model (TSModel) using observational data from city A and subsequently applied it to forecast for city B and C.}
  \label{fig:Map}
\end{figure}

As the field progressed, time series forecasting methods for smart cities have evolved significantly. Vector Autoregression (VAR) models recognize the interdependencies among multiple variables in time series and leverage them to predict future values, but they were limited by their assumption of linearity and their inability to handle non-linear relationships. Autoregressive Integrated Moving Average (ARIMA) models \cite{ARIMA}, built upon the foundation of Autoregressive Moving Average (ARMA) models, address the challenges of non-stationarity encountered by ARMA models. However, ARIMA models can be limited by their strict assumptions about data properties and may face computational inefficiency when applied to large datasets. Recurrent Neural Networks (RNNs) \cite{RNN}, including Long Short-Term Memory (LSTM) \cite{LSTM} and Gated Recurrent Unit (GRU) \cite{GRU}, revolutionized traditional methods by effectively capturing temporal dependencies. However, they encountered challenges in capturing long-range dependencies, were susceptible to vanishing or exploding gradients, and were data hungry~\cite{Vanishing}. These limitations also manifest in location-aware time series forecasting. 
Transformer \cite{Transformer} has emerged as a significant advancement, effectively addressing the limitations of earlier methods by integrating a self-attention mechanism. By capturing both short and long-range dependencies, transformer excels in enhancing forecasting accuracy and surpassing the constraints of previous approaches. While transformer and other RNNs have demonstrated their effectiveness in time series forecasting, they often struggle when confronted with geographic domain shifts where the target urban environments differ significantly from the source urban environments (See Figure \ref{fig:Map}). In such cases, these models tend to have unsatisfactory performance \citep{GDS,AdaptCNNs}. 

Let's delve into an illustrative example concerning urban dynamics to highlight the complexities of predicting urban air pollution, a critical task in time series forecasting. Traditional approaches often fall short by failing to recognize invariant causal relationships between variables, relying instead on spurious correlations that do not hold across different contexts. In this context, $\mathcal{X}$ represents the level of traffic congestion, $\mathcal{Y}$ denotes the air pollution levels, and $\mathcal{Z}$ indicates the prevalence of respiratory illnesses within the same urban area. We establish a direct causal link between traffic congestion ($\mathcal{X}$) and air pollution ($\mathcal{Y}$); as traffic congestion increases, so does air pollution, primarily due to higher vehicle emissions and engine idling. Moreover, there's a consequential relationship between air pollution ($\mathcal{Y}$) and respiratory health issues ($\mathcal{Z}$). Increased exposure to polluted air significantly raises the risk of respiratory ailments, leading to a higher prevalence of such diseases. This interplay illustrates the critical need for air pollution forecasting models that are precise, causally aware, and capable of identifying and leveraging these invariant causal relationships. Our work specifically addresses the intricate dynamics of air pollution within urban landscapes. By focusing on these dynamics, we aim to contribute to the development of more sustainable urban environments. We provide accurate predictions that are instrumental in informing policy decisions and urban planning strategies, ultimately aiming to reduce air pollution and mitigate its adverse health effects. Our approach innovates by integrating advanced causal inference techniques with time series forecasting, offering a novel perspective in the fight against urban air pollution.

However, the correlation between $\mathcal{Y}$ and $\mathcal{Z}$ may appear spurious when considering different urban environments. Some urban areas may have effective pollution control measures, ample green spaces, or favorable air circulation patterns, which mitigate the impact of air pollution on respiratory health. Consequently, the correlation between $\mathcal{Y}$ (level of air pollution) and $\mathcal{Z}$ (prevalence of respiratory illness) may be weakened or even absent in such areas. To improve its accuracy, it is essential to ensure the model considers the underlying causal factors and avoids being misled by spurious correlations. To overcome this limitation, we propose InvarNet, an innovative framework specifically designed to tackle the challenges of OOD generalization in location-aware time series forecasting model for smart cities. InvarNet consists of Invariant LSTM (Invar-LSTM) and Invariant Transformer (Invar-Transformer) models, built upon the LSTM and Transformer architectures, respectively. Integrating the invariant risk minimization (IRM) \cite{IRM} framework enables the models to effectively handle geographic domain shifts, improving their time series forecasting capabilities and ensuring robust performance. In InvarNet, we begin by partitioning the time series based on their respective geographic locations. This partitioning allows us to isolate and analyze the data within specific urban environments. We then train our invariant time series forecasting model with the data from the source environments. Our proposed model is designed to encourage the learning of invariant relationships, rather than relying on spurious correlations that may be present across diverse environments. Therefore, it possesses the ability to generalize effectively within the target geographic domain. The development of InvarNet represents a significant contribution to the field. It provides a solid foundation for further exploration and advancements in OOD location-aware time series forecasting. Through comprehensive evaluations on both synthetic and real-world location-aware time series data, we have demonstrated the effectiveness of our approach, which also opens up a new avenue for research and development in smart cities.
\section{Problem Formulation}
Consider a \textit{multivariate time series} $X=\langle x_1, x_2, ...,x_T \rangle$, where measurement $x_T \in \mathcal{R}^{d}$ is recorded at time step $T$ with $d$ attributes. A set of \textit{location-aware multivariate time series} is denoted as $S = \{(L_{n},X_{n})\}_{n=1}^{N}$, where $N$ is the number of observed locations, $L_{n} = (\text{lat}_n, \text{lon}_n)$ represents the geographic coordinates (latitude and longitude) of a specific location, and $X_n \in \mathcal{R}^{d \times T}$ represents the observed multivariate time series at $n$-th location.\\
\textbf{Problem Statement:} In the context of location-aware multivariate time series, our objective is to develop a mapping function, denoted as $f(\cdot)$, on the training set $S_{tr} \subset S_{all}$. This training set consists of observations $\langle X_{1}, X_{2},...,X_{N} \rangle$ from $N$ locations. The aim is to predict $x_{T+k}$, where $k$ represents the desired number of time steps into the future from the current time step, using the historical data $\langle x_{1}, x_{2},...,x_{T} \rangle$. Furthermore, we aim to ensure that this mapping function exhibits robust performance when applied to other geographic locations within $S_{all}$ that were not included in the training set.

\section{Methodology}
In this section, we present our novel forecasting approaches by developing IRM training algorithms on two deep neural network architectures: Invariant Long Short-Term Memory (\textbf{Invar-LSTM}, see Figure \ref{fig:Invar-LSTM})  and Invariant Transformer (\textbf{Invar-Transformer}, see details in Subsection \ref{invar-trans}). 
\subsection{Invariant Long Short-Term Memory}
First, we briefly introduce the application of LSTM in location-aware time-series forecasting. Given a training set $S_{tr}$, which contains a set of multivariate time series from a variety of locations, our goal is to train an LSTM network $f(\cdot)$ that maps $S_{tr}^{1:T}$ to $S_{tr}^{T+k}$. We optimize the network by minimizing the loss function $\mathcal{L}(S_{tr}^{T+k},\widehat{S}_{tr}^{T+k})$, where $S_{tr}^{T+k}$ denotes the ground truth, $\widehat{S}_{tr}^{T+k}$ is the predicted value. It is often the modeling choice that $\mathcal{L}$ is a convex and differentiable function, such as mean square error and cross-entropy. We observe that directly training the LSTM network in this traditional empirical risk minimization manner leads to poor performance when applied it to other locations that are not included in $S_{tr}$. We therefore explore the new IRM training paradigm for location-aware time-series forecasting.

\begin{figure}[h]
  \centering
  \includegraphics[width=0.45\textwidth]{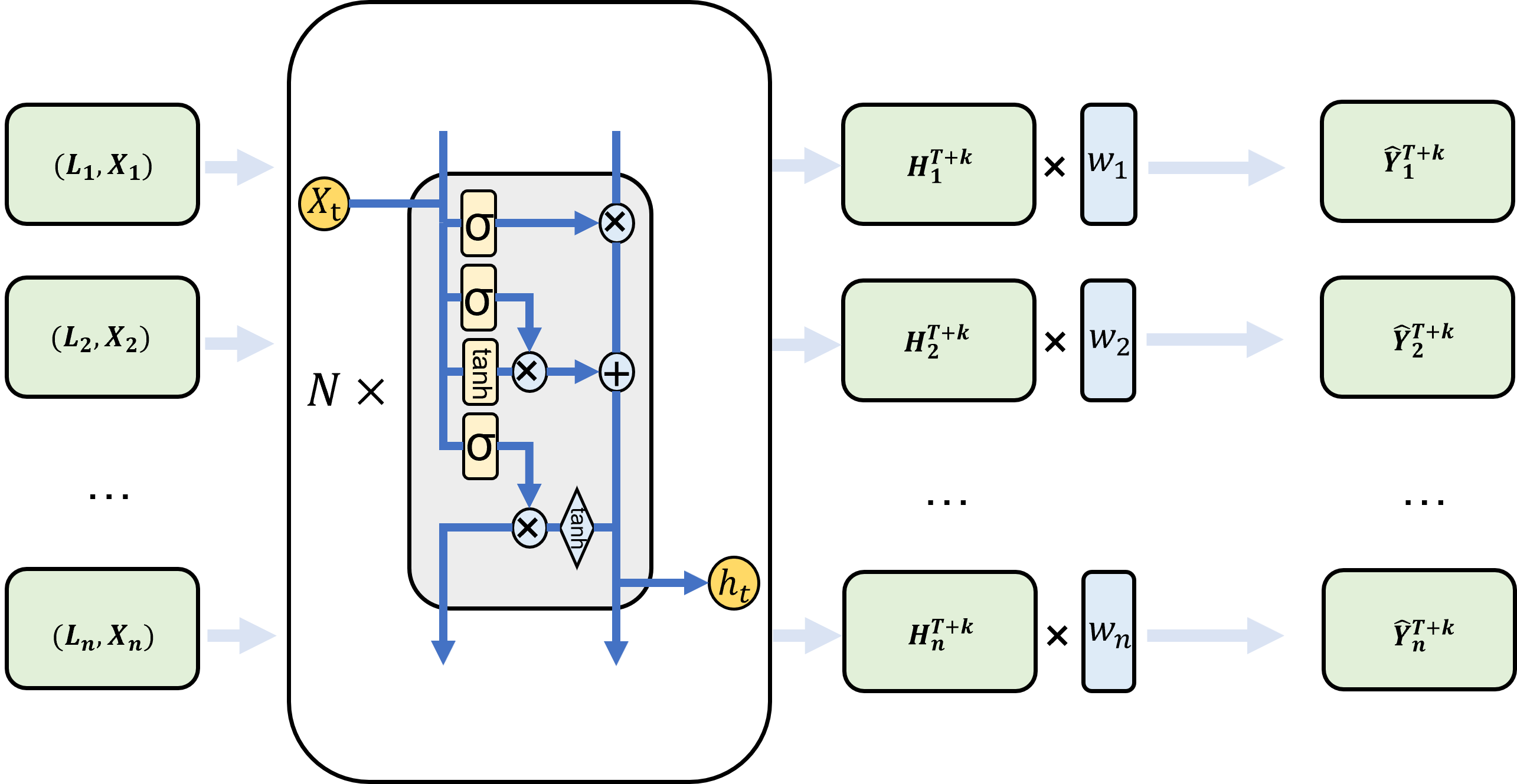}
  \captionsetup{justification=centering}
  \caption{Invar-LSTM for Location-Aware Time Series Forecasting.}
  \label{fig:Invar-LSTM}
\end{figure}


\noindent \textbf{Invar-LSTM} shown in Figure \ref{fig:Invar-LSTM} builds upon the well-established LSTM architecture, integrating it with invariance properties derived from the IRM framework. 
In our approach, we begin by splitting the training set based on the locations from which the data was collected. This results in a set of location-aware multivariate time series pairs, denoted as $(L_{i}, X_{i})_{i=1}^{N}$, where $N$ represents the number of locations. Each pair consists of a geographic location $L$ and a corresponding multivariate time series $X = \langle x_{1}, x_{2},...,x_{T} \rangle$, where $T$ denotes the number of timesteps. Here, $x_{t} \in \mathcal{R}^{d}$ represents a data point at time $t$ with $d$ features.

To capture the temporal dependencies within the time series, we employ the Long Short-Term Memory (LSTM) architecture. At each timestep $t$, the hidden state $h_{t} \in \mathcal{R}^{h}$ is computed using the following equation:
\begin{equation}
    h_{t} = f(W_{input} \cdot x_{t} + W_{hidden} \cdot h_{t-1} + b),
\end{equation}
here, $W_{input} \in \mathcal{R}^{h \times d}$ is the weight matrix for inputs, $W_{hidden} \in \mathcal{R}^{h \times h}$ is the weight matrix for hidden states from the previous timestep, and $b \in \mathcal{R}^{h}$ represents the bias term.

In a traditional LSTM approach, the hidden state $h_{t}$ is typically passed through a dense layer to predict future timesteps, resulting in $Y = f(W_{output} \cdot h_{t})$, where $Y \in \mathcal{R}^{d \times k}$ represents the $d$ features and future $k$ time steps to be predicted. 

To derive more generalizable prediction models, we integrate IRM training scheme to enhance both the prediction accuracy and robustness across different locations by encouraging the time-series forecasting models to learn invariant representations in changing urban environments.

In the case of \textbf{Invar-LSTM}, we introduce an additional step. After obtaining the result from the dense layer, denoted as $H = f(W_{logit} \cdot h_{t})$, where $H \in \mathcal{R}^{d \times k}$ and $W_{logit} \in \mathcal{R}^{d \times k \times h}$, we incorporate an invariant weight matrix $w_{inv}$, which represents an all-\textbf{1} matrix of size $\mathcal{R}^{d \times k}$. The output is calculated as $Y = H \cdot w_{inv}$, where $(\cdot)$ denotes the Hadamard product.
To optimize the model, the objective is defined as follows:
\begin{equation}\label{eq7}
    \min \sum_{e\in\mathcal{E}}\mathcal{R}^{e}(w_{inv}\cdot H) + \lambda\cdot||\nabla_{w_{inv}|w_{inv}=1.0}\mathcal{R}^{e}(w_{inv}\cdot H) ||^{2}_{2}.
\end{equation}

In the equation \eqref{eq7}, $\mathcal{E}$ represents urban environments in the training set, $\mathcal{R}^{e}$ denotes the metric-specific loss function, and $\lambda \in [0,\infty)$ is a regularization parameter. While also penalizing the deviation of the gradients with respect to $w_{inv}$ from their values at $w_{inv}=1.0$.


\subsection{Invariant Transformer}
\label{invar-trans}
We also leverage the capabilities of the transformer architecture. Unlike recurrent models, which face inherent limitations in parallelization, especially when confronted with longer sequence lengths. We mitigate the sequential nature's hindrance, paving the way for enhanced parallelization and improved performance.

In the training of Invar-Transformer, we also incorporate the essential step of dividing the training set based on locations. Once the data is separated, we input the respective subsets into the model. Subsequently, we perform position encoding on the data $x_{t}$ positioned at time $t$ by following the steps outlined below:
$$\textbf{PE}(t)_{i}=\left\{
\begin{aligned}
    &sin(\omega_{i}t) &i\%2=0\\
    &cos(\omega_{i}t) &i\%2=1\\
\end{aligned}
\right.$$
In this context, $\textbf{PE}$ represents the positional encoding, while $\omega_{i}$ signifies the manually designed frequency for each dimension. To seamlessly integrate positional information, we perform the sum of the input embedding and positional embedding, denoted as $X_{in}$. This operation effectively incorporates the positional encoding within the input data. 

To acquire the query matrix $Q$, keys matrix $K$, and values matrix $V$, we utilize three separate weights: $W_{Q}$, $W_{K}$, and $W_{V}$, correspondingly. Each weight is then multiplied by $X_{in}$ to generate the respective matrices.

In the self-attention module, after obtaining $Q$, $K$, and $V$ from the input time series $X_{in}$, the computation involves a scaled dot product self-attention mechanism. This mechanism can be mathematically expressed as follows:
\begin{equation}
    z = \textsf{Attention}(Q,K,V) = \textsf{softmax}(\frac{Q K^{T}}{\sqrt{d^{k}}})V.
\end{equation}
where $d^{k}$ is the dimension of keys.

The Transformer utilizes multi-head attention (\textsf{M-H Attention}) with $M$ distinct sets of learned projections instead of a single attention function. This approach can be represented as:
\begin{equation}
    \textsf{M-H Attention}(Q,K,V)= \textsf{Concat}(\textsf{Attn}_{1},...,\textsf{Attn}_{M})W^{o}
\end{equation}
Here, $\textsf{Attn}_{i} = \textsf{Attention}(Q_{i},K_{i},V_{i})$ represents the $i^{th}$ self-attention module, and $W^{o}$ denotes the output weight matrix. Notably, the dimension of the output $z$ from the \textsf{Multi-Head Attention} module remains the same as the input $X_{in}$. The value of $z$ is subsequently fed through the position-wise feed-forward layer, which comprises two linear transformations with a ReLU activation applied in between. This mathematical representation can be expressed as:
\begin{equation}
    F(z) = \textsf{ReLU}(zW_{1} + b_{1})W_{2} + b_{2}.
\end{equation}
Moreover, it is important to note that the dimension of $F(z)$ remains consistent with the input $X_{in}$. Subsequently, we proceed to map the input $F(z)$ to the output $H \in \mathcal{R}^{d \times k}$ for each location. In the context of the Invar-Transformer, after obtaining $H \in \mathcal{R}^{d \times k}$, we incorporate an invariant weight matrix $w_{inv}$ that is an all-\textbf{1} matrix of size $\mathcal{R}^{d \times k}$, similar to the approach used in LSTM. The output is calculated as $Y = H \cdot w_{inv}$. To train the transformer model, we optimize it using the same objective function (Equation \ref{eq7}) as used in InvarLSTM.

\section{Experiments}
\subsection{Datasets}
\textbf{Synthetic Data:} 
We first evaluate our methods using synthetic data, employing a \textit{structural equation model} represented as follows:
\begin{equation}\label{eq10}
\begin{gathered}
   \mathcal{X}_{t} \leftarrow \mathcal{X}_{t-1} + \mathcal{N}(0,\sigma^{2}) \\
   \mathcal{Y}_{t} \leftarrow \mathcal{Y}_{t-1} + \mathcal{X}_{t-1} + \mathcal{N}(0,\sigma^{2})\\
   \mathcal{Z}_{t} \leftarrow \mathcal{Z}_{t-1} + \mathcal{Y}_{t-1} + \mathcal{N}(0,1)
\end{gathered}
\end{equation}
Here, we represent the temporal dependencies among variables as recursive equations. Specifically, $\mathcal{X}_{t}$ is influenced by its previous value, $\mathcal{Y}_{t}$ depends on its previous value as well as $\mathcal{X}_{t-1}$, and $\mathcal{Z}_{t}$ is influenced by its previous value, $\mathcal{Y}_{t-1}$, and a normally distributed noise term with zero mean and unit variance.
Moreover, in the urban environment represented by $e \in \mathcal{E}_{tr}$, we consider that the value of $\sigma^{2}$ varies with different urban environments.
\begin{figure}[h]
  \centering
  \includegraphics[width=1.0\linewidth]{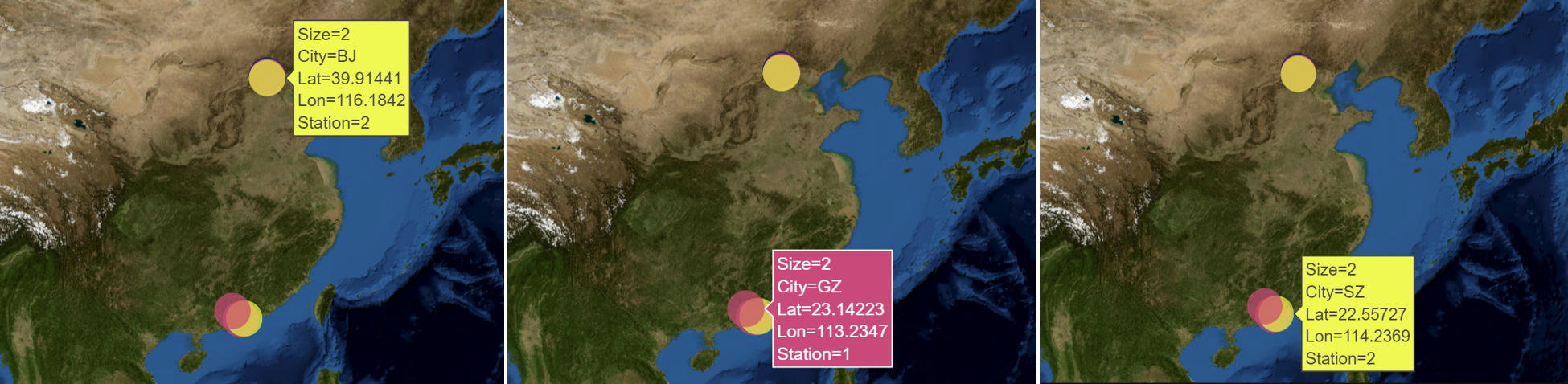}
  \captionsetup{justification=centering}
  \caption{Visualization of the Geographic Distribution of Cities.}
  \label{fig:VisGeo-1}
\end{figure}

\noindent \textbf{Real-world Data:} We conducted our experiments using a real-world dataset \cite{RWdata} consisting of air quality measurements collected in Beijing (BJ), Shenzhen (SZ), and Guangzhou (GZ), China, which are visualized in the Figure \ref{fig:VisGeo-1}. The dataset covers the period from May 1, 2014, to April 30, 2015. Within this real-world air quality dataset, we have six attributes that are measured: $PM_{2.5}$, $PM_{10}$, $NO_{2}$, $CO$, $O_{3}$, and $SO_{2}$. These attributes are monitored in multiple stations where Beijing has a total of 36 stations, Guangzhou has 30 stations, and Shenzhen has 10 stations. Each station is associated with precise geographical coordinates in terms of latitude and longitude. This information enables accurate spatial analysis of the air quality data. The dataset provides a high temporal resolution, with hourly measurements of pollutant concentrations. This level of granularity allows for detailed analysis of temporal variations, facilitating the identification of daily fluctuations and seasonal trends in air pollution.
\subsection{Evaluation Metrics}
Represent the predicted results and the ground truth in the testing set across $N$ geographic locations as $\widehat{Y}^{T+k}_{n}$ and $Y^{T+k}_{n}$, respectively. The error metrics can be defined as follows:
\begin{equation}
    \textsf{Mean Absolute Error(MAE)} = \frac{1}{|N|}\sum_{n=1}^{|N|}|Y^{T+k}_{n}-\widehat{Y}^{T+k}_{n}|.
\end{equation}
\begin{equation}
    \textsf{Mean Squared Error(MSE)} = \frac{1}{|N|}\sum_{n=1}^{|N|}(Y^{T+k}_{n}-\widehat{Y}^{T+k}_{n})^2.
\end{equation}
For the \textsf{MAE} and \textsf{MSE}, lower values indicate better accuracy and prediction performance.

\subsection{Main Results}
\textbf{Results of Synthetic Data Analysis:} We conducted our experiments initially on synthetic data. The purpose was to showcase the superiority of the Invariance-based Time Series Forecasting Model~(Invar-TSModel) in learning invariant representations, in comparison to the results obtained with the baseline TSModel trained using empirical risk minimization.

\begin{figure}[h]
  \centering
  \includegraphics[width=0.8\linewidth]{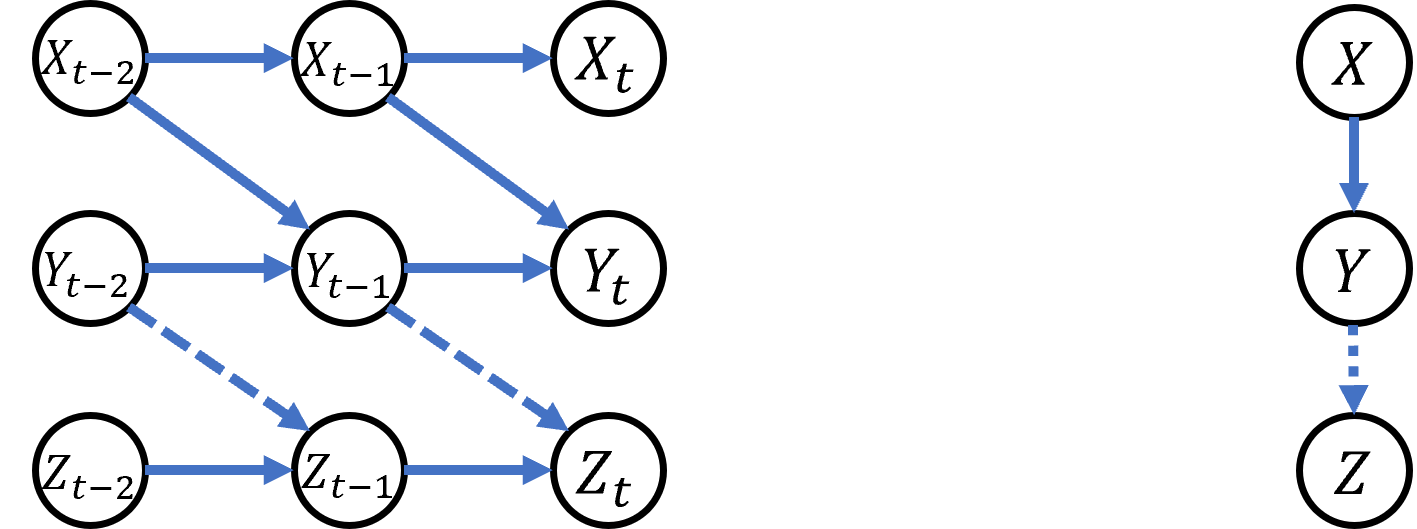}
  \captionsetup{justification=centering}
  \caption{Temporal Invariance (Left Solid Line) and Spurious (Left Dotted Line) Relationship, Spatial Invariance (Right Solid Line) and Spurious (Right Dotted Line) Relationship.}
  \label{fig:STInv}
\end{figure}

In our synthetic time series data as shown in Equation \eqref{eq10}, there are two distinct invariance relationships: The first invariance relationship, denoted as $\mathcal{X}_{t-1} \rightarrow \mathcal{X}_{t}$, originates from a temporal perspective. This invariance relationship is primarily driven by temporal dependence, where the historical data $\mathcal{X}_{t-1}$ serves as the causal factor, influencing the data observed at the current time step, $\mathcal{X}_{t}$, as the effect. The second invariance relationship is from a spatial perspective, where the attribute $\mathcal{X}$ serves as the cause and $\mathcal{Y}$ as the effect. 
Additionally, it is worth noting that the relationship between $\mathcal{Y}$ and $\mathcal{Z}$ exhibits invariance only in a specific urban environment, we refer to this as spurious correlation in other urban environments. In our synthetic time series data, the invariant causal relationship from $\mathcal{X}$ to $\mathcal{Y}$ remains invariant and unaffected by changes in the environments $\sigma^2$. However, the relationship between $\mathcal{Y}$ and $\mathcal{Z}$ exists only when $\sigma^2=1$, indicating that for other urban environments where $\sigma^2 \neq 1$, this invariant relationship breaks down and is considered a spurious correlation. In such instances, the causal structure can be depicted as shown in Figure \ref{fig:STInv}. As a result, when attempting to predict $\mathcal{Y}$ using both $\mathcal{X}$ and $\mathcal{Z}$, the presence of the spurious correlation between $\mathcal{Z}$ and $\mathcal{Y}$ can weaken the performance of the model. 

To be more specific, given the data in Equation (\ref{eq10}), for the model $\mathcal{Y}=\alpha_{1}\mathcal{X},$ the optimal solution is $\alpha_{1}^{*}=1$.
Similarly, for the model $\mathcal{Y}=\alpha_{2}\mathcal{Z},$ the optimal solution is $\alpha_{2}^{*}=\frac{\sigma^{2}}{\sigma^{2}+0.5}$. When utilizing both $\mathcal{X}$ and $\mathcal{Z}$ to predict $\mathcal{Y}$ with the model $\mathcal{Y}=\alpha_{1}\mathcal{X} + \alpha_{2}\mathcal{Z},$ the optimal solutions for $\alpha_{1}$ and $\alpha_{2}$ are found to be $\alpha_{1}^{*}=\frac{1}{\sigma^{2}+1}$ and $\alpha_{2}^{*}=\frac{\sigma^{2}}{\sigma^{2}+1}$, respectively. We have provided the derivation in the \textbf{Appendix \ref{sec:Appendix}}. These results highlight the impact of the varying urban environments $\sigma^{2}$ on the coefficients when predicting $\mathcal{Y}$. In other words, when using $\mathcal{X}$ alone to predict $\mathcal{Y}$, the coefficient remains unaffected by changes in the urban environment. However, when incorporating $\mathcal{Z}$ (either alone or in combination with other invariant variables) to predict $\mathcal{Y}$, the coefficient is influenced by the varying urban environment $\sigma^{2}$.

This theoretical analysis highlights a crucial insight: certain time-series forecasting models may struggle to generalize effectively due to their failure to capture the invariant relationship between $\mathcal{X}$ and $\mathcal{Y}$. Instead, these models tend to place greater emphasis on the spurious correlation between $\mathcal{Z}$ and $\mathcal{Y}$ occurs in specific urban environments, limiting their generalizability. 
\begin{table}[h]
\centering
\captionsetup{justification=centering}
\caption{Evaluation Results on Synthetic Data.}
\label{tab:syn}
\resizebox{8.0cm}{!}{
\begin{tabular}{lccc}
    \toprule
    Methods & \multicolumn{3}{c}{Env-Type} \\
    \cmidrule{2-4}
     & 2 & 3-1B & 3-2G \\
    \midrule
    MSE (TSModel) & 3.614$\pm$0.002 & 4.735$\pm$0.002 & 1.425$\pm$0.002 \\
    MSE (Invar-TSModel) & 2.327$\pm$0.002 & 2.992$\pm$0.001 & 1.155$\pm$0.002 \\
    \addlinespace
    MAE (TSModel) & 1.513$\pm$0.003 & 1.735$\pm$0.003 & 0.955$\pm$0.001 \\
    MAE (Invar-TSModel) & 1.214$\pm$0.001 & 1.376$\pm$0.001 & 0.863$\pm$0.002 \\
    \bottomrule
\end{tabular}}
\end{table}

\begin{figure}[h]
  \centering
  \includegraphics[width=\linewidth]{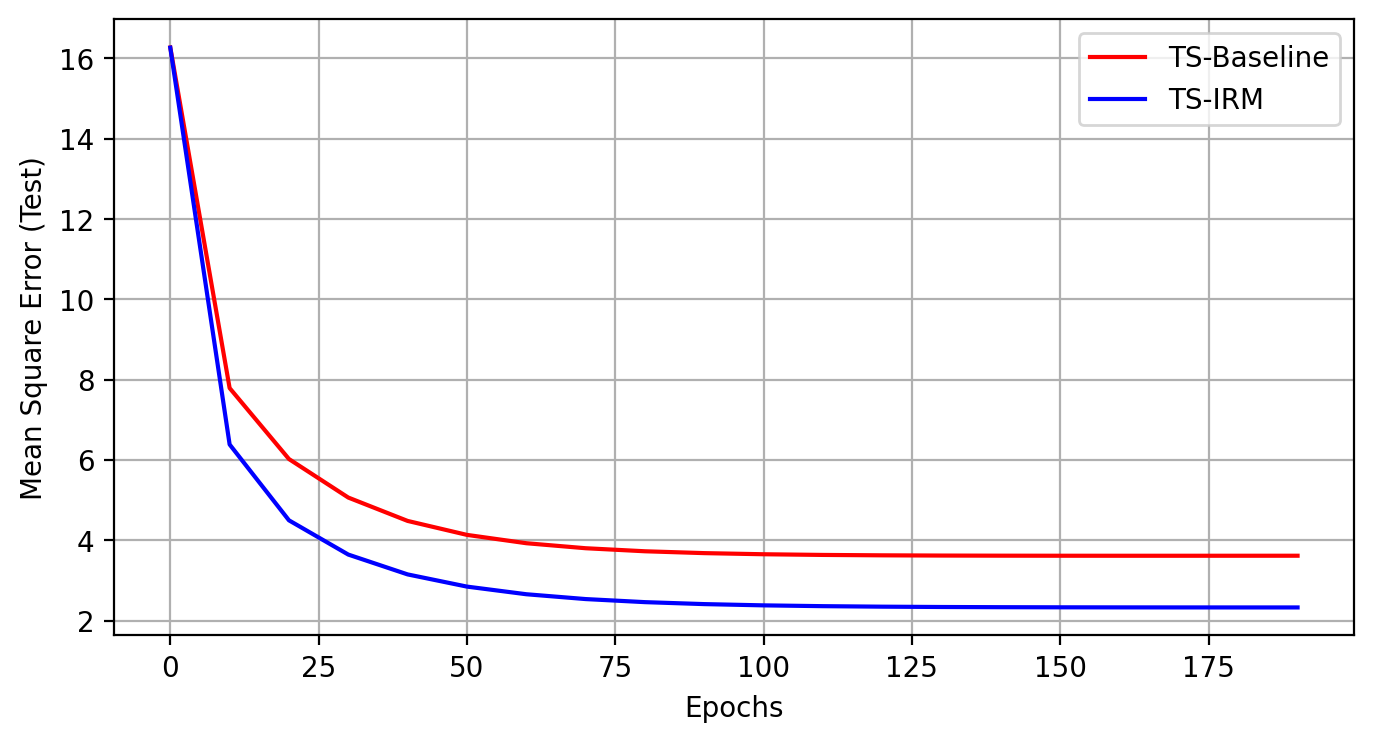}
  \captionsetup{justification=centering}
  \caption{Test Error Variation in Env-Type=2 Setting.}
  \label{fig:TrainCurve}
\end{figure}

To validate this idea, we conducted a comparative analysis of the performance between a traditional TSModel and its Invariance-based counterpart on the synthetic data. The results of this comparison are presented in Table \ref{tab:syn}. The training environments included three distinct types. For the first type of environment (Env-Type=2), with a total of two environments, in our experiments, we set $\mathcal{E}_{train} = \{\sigma^{2}_{e_{1}}=0.1, \sigma^{2}_{e_{2}}=1.0\}$ and $\mathcal{E}_{test}=\{\sigma^{2}_{e_{t}}=2.0\}$. Notably, our proposed Invar-TSModel showcased a better forecasting performance when compared to the traditional TSModel under this setting (See Figure \ref{fig:TrainCurve}).
Moving on to the second type of environment (Env-Type=3-1B), we introduced an additional data to the training set that exhibited a larger domain shift compared to the target environments. Here, we established $\mathcal{E}_{train}=\{\sigma^{2}_{e_{1}}=0.1, \sigma^{2}_{e_{2}}=1.0, \sigma^{2}_{e_{3}}=0.01\}$ and $\mathcal{E}_{test}=\{\sigma^{2}_{e_{t}}=2.0\}$. In this scenario, the traditional TSModel experienced a decline in performance, while the Invar-TSModel demonstrated relatively better robustness in the face of domain shifts, although its performance was not as strong as in the previous setting. Finally, in the third setting (Env-Type=3-2G), we added data specifically from the target environment. The performance of the traditional model witnessed an improvement. Under these conditions, with $\mathcal{E}_{train} = \{\sigma^{2}_{e_{1}}=0.1, \sigma^{2}_{e_{2}}=1.0, \sigma^{2}_{e_{3}}=2.0\}$ and $\mathcal{E}_{test} = \{\sigma^{2}_{e_{t}}=2.0\}$, our proposed Invar-TSModel continued to outperform the traditional TSModel.

These findings on synthetic data illustrate the advantageous capabilities of the Invar-TSModel in effectively handling domain shifts caused by varying training urban environments, as it consistently outperforms the traditional TSModel in diverse settings.

\begin{figure}[ht]
  \centering
  \includegraphics[width=\linewidth]{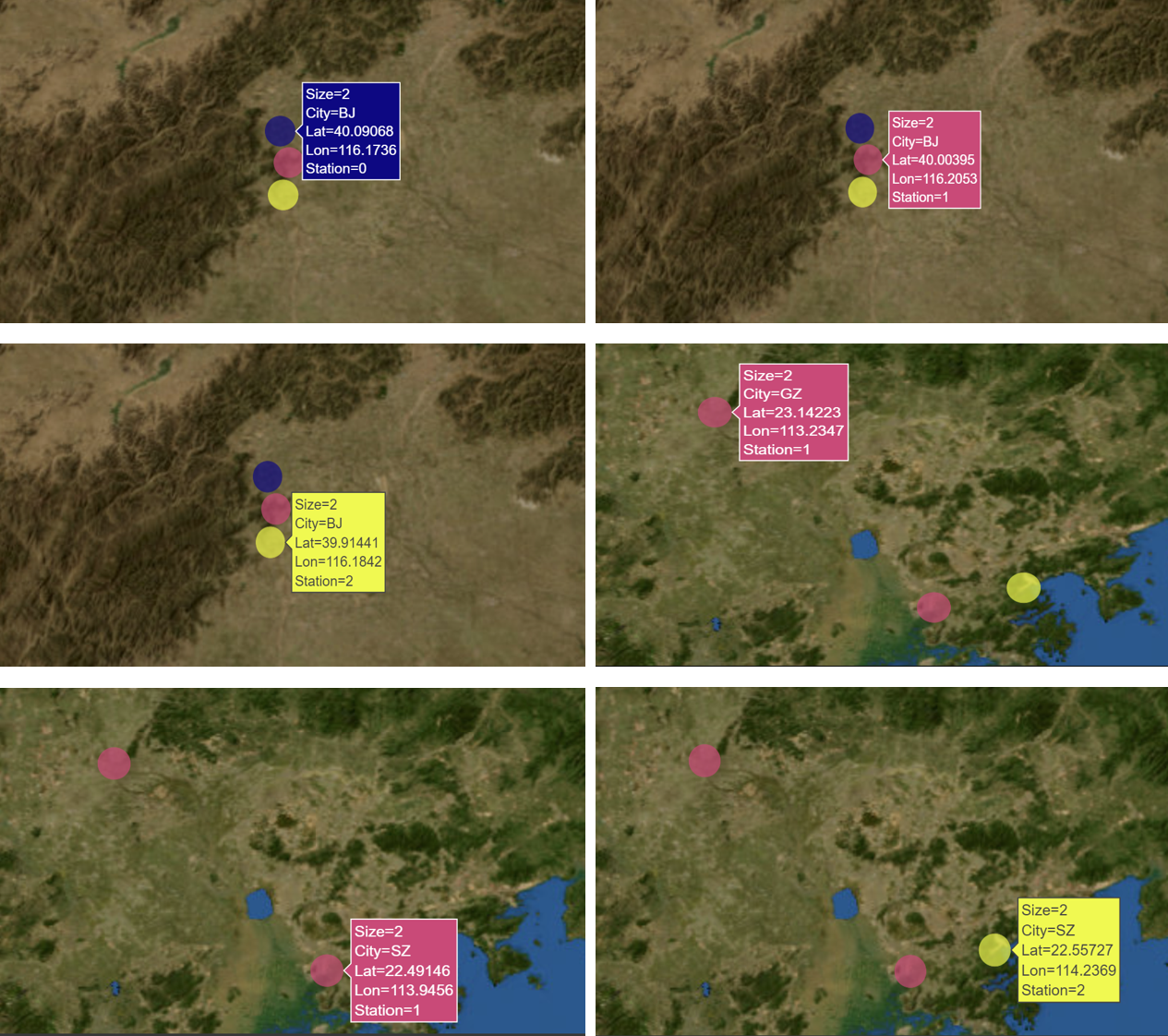}
  \captionsetup{justification=centering}
  \caption{Geographic Distribution of Monitoring Stations used in Our Experiments.}
  \label{fig:Vis-Stations}
\end{figure}

\noindent\textbf{Results of Real-World Data Analysis:}
Our experiments on real-world data were designed with three different settings. Firstly, we selected a station from Beijing, denoted it as BJ-0, to serve as the test environment for all three settings. Then, in the first setting, we chose two different stations from Beijing, denoted them as BJ-1 and BJ-2, as the training environments. Moving on to the second setting, we selected two different stations from Shenzhen, namely SZ-1 and SZ-2, as our training environments. In the third setting, we included SZ-1 as part of the training environment, but instead of SZ-2, we replaced it with a station from Guangzhou called GZ-1.

For all three settings, we utilized historical data from the previous seven days to predict values for the following three days. We trained LSTM, Invar-LSTM, Transformer, and Invar-Transformer models on the respective training environments for each setting. Except for the invariance mechanism, the backbone structure for LSTM and Transformer models remained the same. The performance of these models was then evaluated on the same BJ-0 test environment. The results of these evaluations can be found in Table \ref{tab:realworld}.

In our analysis of the results, we observed distinct patterns across the different settings. In the first setting, where there was a small geographic domain shift within Beijing, we found that Invariance-based TSModel showed a slight improvement compared to the traditional TSModel. This suggests that the invariance mechanism helps mitigate the effects of geographic shifts, albeit to a limited extent. Moving to the second setting, which involved a larger geographic domain shift from Shenzhen to Beijing according to a significant increase in mean square error generated by the model. However, Invar-LSTM exhibited a notable improvement in performance compared to the first setting. This improvement can be attributed to the inherent ability of Invar-LSTM to learn invariant representations, which proves beneficial in the face of larger geographic domain shifts. In the third setting, we also tested our model with training data not only from one city but also from various other cities, where we replaced SZ-2 with GZ-1, LSTM performed better compared to the second setting. This indicates that the distribution of time series data from GZ-1 is more similar to BJ-0 in comparison to SZ-2. In this scenario, Invar-LSTM showcased its capability to handle the invariant relationship that arises in more diverse urban environments, resulting in better performance than using LSTM alone.

\begin{table}[h]
\centering
\captionsetup{justification=centering}
\caption{Evaluation Results on Real-world Data (Metric: Mean Square Error).}
\resizebox{8.0cm}{!}{
\begin{tabular}{lccc}
\toprule
Method / Env-Type & 2BJ-1BJ & 2SZ-1BJ & (SZ+GZ)-1BJ \\
\midrule
LSTM & 0.655$\pm$0.002 & 0.986$\pm$0.002 & 0.959$\pm$0.002 \\
Invar-LSTM & 0.652$\pm$0.002 & 0.942$\pm$0.011 & 0.896$\pm$0.025 \\
Transformer & 0.238$\pm$0.015 & 0.860$\pm$0.005 & 0.819$\pm$0.005 \\
Invar-Transformer & 0.228$\pm$0.017 & 0.785$\pm$0.002 & 0.654$\pm$0.005 \\
\bottomrule
\end{tabular}}

\label{tab:realworld}
\end{table}

Additionally, we extended our analysis to include popular Transformer models by integrating the invariance mechanism. We observed that the combined approach exhibited better performance than LSTM. By leveraging the advantages of the Transformer architecture and incorporating the invariance mechanism, we achieved the best performance in addressing geographic domain shifts in location-aware time series forecasting problems.
\section{Discussion}
Our research was motivated by the causal invariance concept to address the location-aware time series forecasting out-of-distribution problem in smart cities. We started by exploring spatio-temporal and causal analysis, recognizing that certain time series models excel in capturing temporal invariance. However, many time series models overlook this crucial aspect on the spatial domain, leading to poor performance in such situations. Some models have proposed changes to the inner architecture to enhance spatial considerations but often at the cost of increased computational complexity. In contrast, we aimed to develop an efficient and compact model that could handle both spatial and temporal invariant relationships, thereby improving robustness and accuracy.

Our experiments revealed the prevalence of geographic domain shifts within the data, even within a single city. Another challenge arose from uneven training data volume, where, for example, City 1 had 10,000 data points while City 2 had only 2,000. Such data imbalance could cause models to favor fitting to City 1, resulting in poor generalization to other locations. Leveraging the geo-location aspect in combination with time series data, we devised techniques to encourage the model to identify and adapt to invariant relationships present in each environment, despite the changing conditions.

\section{Conclusion}
In conclusion, this study introduced and evaluated a novel method for city-level time series forecasting with the InvarNet framework, offers a more promising and effective approach to address the geographic domain shifts caused by changing urban environments. By incorporating both spatial and temporal invariance into LSTM and transformer architectures, we successfully navigated the complexities posed by geographic domain shifts. Our experiments, conducted on both synthetic and real-world time-series data in urban areas, showcased the superior performance and robustness of InvarNet compared to traditional time-series forecasting models. The choice of LSTM and Transformer as our base models was driven by our primary focus on forecasting time series data. Moreover, this methodology has the potential to be expanded to include regression models such as linear regression and geographically weighted regression \cite{GWR} in future work. These results hold promising prospects for applications across diverse domains, such as utilizing InvarNet for climate modeling to forecast future climatic conditions and employing our proposed models in urban planning to anticipate traffic patterns.

\bibliographystyle{ACM-Reference-Format}
\bibliography{sample-base}


\appendix
\section{Appendix}

\label{sec:Appendix}
To understand the impact of changing geographic environments $\sigma^2$ on the performance of the model, we first assumed a specific data generation process that follows the \textit{structural equation model} outlined below:
\begin{equation}\label{eq13}
\begin{gathered}
   \mathcal{X} \leftarrow \epsilon_{1} \\
   \mathcal{Y} \leftarrow \mathcal{X} + \epsilon_{2}\\
   \mathcal{Z} \leftarrow \mathcal{Y} + \epsilon_{3}
\end{gathered}
\end{equation}
where $\epsilon_{1}, \epsilon_{2} \stackrel{i.i.d}{\sim} \mathcal{N}(0,\sigma^{2})$, and $\epsilon_{3} \sim \mathcal{N}(0,1)$. In different geographic environments, $\sigma^2$ will change with the environments. Next, we will provide empirical evidence to demonstrate the impact of changing geographic environments on the predictive performance of the model. 

Given a model for predicting $\mathcal{Y}$ using variables $\mathcal{X}$ and $\mathcal{Z}$, where $\mathcal{X}$, $\mathcal{Y}$, and $\mathcal{Z}$ follow the \textit{structural equation model} as shown in equation \eqref{eq13}:
$$\widehat{\mathcal{Y}} = \alpha_{1} \mathcal{X} + \alpha_{2} \mathcal{Z}.$$

Firstly, we assume that our objective is to find an optimal value for $\alpha_{1}$ such that the estimator $f(\mathcal{X})=\alpha_{1}\mathcal{X}$ provides a reliable approximation of the variable $\mathcal{Y}$. In this case, our objective function can be defined as follows:
$$\min_{\alpha_{1}}\mathbb{E}_{\epsilon}[\mathcal{Y}-f(\mathcal{X})]^2.$$
Denoted  $\mathbb{E}_{\epsilon}(\mathcal{Y}-f(\mathcal{X}))^2$ as $F(\alpha_{1})$; thus
\begin{align*}
     F(\alpha_{1}) &= \mathbb{E}_{\epsilon_{1},\epsilon_{2}}[\mathcal{X} +\epsilon_{2}-\alpha_{1} \mathcal{X}]^{2}\\
     &= \mathbb{E}_{\epsilon_{1},\epsilon_{2}}[(1-\alpha_{1})\epsilon_{1} + \epsilon_{2}]^{2}\\
     &= \mathbb{E}_{\epsilon_{1},\epsilon_{2}}[(1-\alpha_{1})^2\epsilon_{1}^{2} + 2(1-\alpha_{1})\epsilon_{1}\epsilon_{2}+\epsilon_{2}^{2}]\\
     &= (1-\alpha_{1})^2\mathbb{E}[\epsilon_{1}^{2}] + 2(1-\alpha_{1})\mathbb{E}[\epsilon_{1}\epsilon_{2}] + \mathbb{E}[\epsilon_{2}^{2}]\\
     &= (1-\alpha_{1})^{2}\sigma^2 + \sigma^2.
\end{align*}
$\epsilon_{1}$ and $\epsilon_{2}$ are independent random variables, so $$\mathbb{E}[\epsilon_{1}\epsilon_{2}]=\mathbb{E}[\epsilon_{1}]\mathbb{E}[\epsilon_2]=0,$$
the objective function will be transformed into:
$$\min\limits_{\alpha_{1}} F(\alpha_{1}) = \min_{\alpha_{1}} (1-\alpha)^2\sigma^{2} + \sigma^{2},$$
the optimal solution is $\alpha_{1}^{*}=1$.

Then, we assume that our objective is to find $\alpha_{2}$ s.t. $f(\mathcal{Z})=\alpha_{2} \mathcal{Z}$ is a good estimator of $\mathcal{Y}$, the objective function will be:
$$\min_{\alpha_{2}}\mathbb{E}_{\epsilon}[\mathcal{Y}-f(\mathcal{Z})]^2.$$
Denote $\mathbb{E}_{\epsilon}[\mathcal{Y}-f(\mathcal{Z})]^2$ as $F(\alpha_{2})$; thus,
\begin{align*}
    F(\alpha_{2}) &= \mathbb{E}_{\epsilon_{1},\epsilon_{2},\epsilon_{3}}[(\alpha_{2}\mathcal{Z} -\mathcal{X} - \epsilon_{2})^{2}]\\
    &= \mathbb{E}_{\epsilon_{1},\epsilon_{2},\epsilon_{3}}[(\alpha_{2}(\mathcal{Y}+\epsilon_{3}) -\epsilon_{1} - \epsilon_{2})^{2}]\\
    &= \mathbb{E}_{\epsilon_{1},\epsilon_{2},\epsilon_{3}}[(\alpha_{2}(\epsilon_{1}+\epsilon_{2}+\epsilon_{3}) -(\epsilon_{1} + \epsilon_{2}))^{2}].
\end{align*}
Here, denote $\epsilon_{4} = (\epsilon_{1} + \epsilon_{2}) \sim \mathcal{N}(0,2\sigma^2)$, and thus:
\begin{align*}
    F(\alpha_{2}) &= \mathbb{E}_{\epsilon_{3},\epsilon_{4}}[(\alpha_{2}\epsilon_{3} + \alpha_{2}\epsilon_{4} - \epsilon_{4})^2]\\
    &= \mathbb{E}_{\epsilon_{3},\epsilon_{4}}[(\alpha_{2}\epsilon_{3} + (\alpha_{2}-1)\epsilon_{4})^2]\\
    &= \mathbb{E}_{\epsilon_{3},\epsilon_{4}}[\alpha_{2}^{2}\epsilon_{3}^2 + 2\alpha_{2}\epsilon_{3}(\alpha_{2}-1)\epsilon_{4} + (\alpha_{2}-1)^{2}\epsilon_{4}^{2})]\\
    &= \alpha_{2}^{2}\mathbb{E}[{\epsilon}_{3}^{2}] + 2\alpha_{2}(\alpha_{2}-1)\mathbb{E}[{\epsilon}_{3}{\epsilon}_{4}] + (\alpha_{2}-1)^{2}\mathbb{E}[{\epsilon}_{4}^{2}]\\
    &= (\alpha_{2}-1)^{2} \cdot 2\alpha_{2}^{2} + \alpha_{2}^{2},
\end{align*}
$\epsilon_{3}$ is a random variable, which independent of $\epsilon_{4}$, so $$\mathbb{E}[\epsilon_{3}\epsilon_{4}]=\mathbb{E}[\epsilon_{3}]\mathbb{E}[\epsilon_4]=0,$$
meanwhile, the objective function is:
$$\min\limits_{\alpha_{2}} F(\alpha_{2}) = \min\limits_{\alpha_{2}} (\alpha_{2}-1)^{2} \cdot 2\alpha_{2}^{2} + \alpha_{2}^{2},$$
the optimal solution for the objective function is $\alpha_{2}^{*} = \frac{\sigma^2}{\sigma^2 + 0.5}$.

At last, we assume that our objective is to find both $\alpha_{1}$ and $\alpha_{2}$ s.t. $f(\mathcal{X},\mathcal{Z})=\alpha_{1} \mathcal{X} + \alpha_{2} \mathcal{Z}$ is a good estimator of $\mathcal{Y}$, the objective function will be:
$$\min_{\alpha_{1},\alpha_{2}}\mathbb{E}_{\epsilon}[\mathcal{Y}-f(\mathcal{X},\mathcal{Z})]^2.$$
Denote $\mathbb{E}_{\epsilon}[\mathcal{Y}-f(\mathcal{X},\mathcal{Z})]^2$ as $F(\alpha_{1},\alpha_{2})$; thus,
\begin{align*}
    F(\alpha_{1},\alpha_{2}) &= \mathbb{E}[(\alpha_{1}\mathcal{X} + \alpha_{2}\mathcal{Z} - (\mathcal{X} + \epsilon_{2}))^2]\\
    &= \mathbb{E}[((\alpha_{1} - 1)\mathcal{X} + \alpha_{2}(\mathcal{X} + \epsilon_{2} + \epsilon_{3}) - \epsilon_{2})^2]\\
    &= \mathbb{E}[((\alpha_{1} + \alpha_{2}-1)\epsilon_{1} + (\alpha_{2}-1)\epsilon_{2} + \alpha_{2}\epsilon_{3})^2]\\
    &= \mathbb{E}[(\alpha_{1} + \alpha_{2}-1)^2\epsilon_{1}^{2} + (\alpha_{2}-1)\epsilon_{2}^{2} + \alpha_{2}\epsilon_{3}^{2}\\
    &+ 2(\alpha_{1} + \alpha_{2}-1)(\alpha_{2}-1)\epsilon_{1}\epsilon_{2} + 2(\alpha_{1} + \alpha_{2}-1)\alpha_{2}\epsilon_{1}\epsilon_{3}\\
    &+ 2(\alpha_{2}-1)\alpha_{2}\epsilon_{2}\epsilon_{3}]\\
    &= (\alpha_{1} + \alpha_{2}-1)^2\mathbb{E}[\epsilon_{1}^2] + (\alpha_{2}-1)^2\mathbb{E}[\epsilon_{2}^2] + \alpha_{2}^2\mathbb{E}[\epsilon_{3}^{2}]\\
    &+ 2(\alpha_{1}+\alpha_{2}-1)(\alpha_{2}-1)\mathbb{E}[\epsilon_{1}\epsilon_{2}]\\ 
    &+ 2(\alpha_{1}+\alpha_{2}-1)\alpha_{2}\mathbb{E}[\epsilon_{1}\epsilon_{3}] + 2(\alpha_{2}-1)\alpha_{2}\mathbb{E}[\epsilon_{2}\epsilon_{3}]\\
    &= ((\alpha_{1} + \alpha_{2} - 1)^2 + (\alpha_2 - 1)^2)\sigma^2 + \alpha_{2}^2,
\end{align*}
$\epsilon_{1}$, $\epsilon_{2}$, and $\epsilon_{3}$ are independent random variables, so $$\mathbb{E}[\epsilon_{1}\epsilon_{2}]=\mathbb{E}[\epsilon_{1}]\mathbb{E}[\epsilon_2]=0,$$
$$\mathbb{E}[\epsilon_{1}\epsilon_{3}]=\mathbb{E}[\epsilon_{1}]\mathbb{E}[\epsilon_3]=0,$$
$$\mathbb{E}[\epsilon_{2}\epsilon_{3}]=\mathbb{E}[\epsilon_{2}]\mathbb{E}[\epsilon_3]=0,$$
and,
$$\mathbb{E}[\epsilon_{1}^{2}]=\sigma^{2},$$
$$\mathbb{E}[\epsilon_{2}^{2}]=\sigma^{2},$$
$$\mathbb{E}[\epsilon_{3}^{2}]=1,$$
we want to find $\alpha_{1}, \alpha_{2}$ to minimize the function $F(\alpha_{1}, \alpha_{2})$, thus:
$$ \frac{\partial{F}}{\partial_{\alpha_{1}}} = 2(\alpha_{1} + \alpha_{2} - 1)\sigma^{2} = 0,$$
$$ \frac{\partial{F}}{\partial_{\alpha_{2}}} = [2(\alpha_{1} + \alpha_{2}-1)+2(\alpha_{2}-1)]\sigma^2 + 2\alpha_{2} = 0,$$
the optimal solution for the objective functions are $\alpha_{1}^{*} = \frac{1}{\sigma^2+1}$ and $\alpha_{2}^{*} = \frac{\sigma^2}{\sigma^2+1}$.
\end{document}